# Evaluation of Genotypic Diversity Measurements Exploited in Real-Coded Representation

Guillaume Corriveau[a,1], Raynald Guilbault[b], Antoine Tahan[b], and Robert Sabourin[c]

[a]*Product Development Engineering - Aerospace, Bombardier Inc., 2351 Alfred-Nobel, Saint-Laurent, Canada, H4N 2M3.*
[b]*Department of Mechanical Engineering, Ecole de technologie superieure, 1100 rue Notre-Dame Ouest, Montreal, Canada  H3C 1K3.*
[c]*Department of Automated Manufacturing Engineering, Ecole de technologie superieure, 1100 rue Notre-Dame Ouest, Montreal, Canada  H3C 1K3.*

**A B S T R A C T**

Numerous genotypic diversity measures (GDMs) are available in the literature to assess the convergence status of an evolutionary algorithm (EA) or describe its search behavior. In a recent study, the authors of this paper drew attention to the need for a GDM validation framework. In response, this study proposes three requirements (monotonicity in individual varieties, twinning, and monotonicity in distance) that can clearly portray any GDMs. These diversity requirements are analysed by means of controlled population arrangements. In this paper four GDMs are evaluated with the proposed validation framework. The results confirm that properly evaluating population diversity is a rather difficult task, as none of the analysed GDMs complies with all the diversity requirements.

*Keywords:* Diversity measures, evolutionary algorithms, premature convergence.

## 1.  Introduction

One of the major problems with evolutionary algorithms (EAs) is premature convergence towards a suboptimal solution (De Jong, 1975; Mauldin, 1984; Goldberg, 1989; Eshelman & Schaffer, 1991). This is due to a lack of diversity within the population. Single-site convergence schemes often lead to diversity losses, while strategies favoring multi-site convergence are considered to ensure better diversity. Among the most popular ways to solve this problem are the promotion of diversity approaches (Matsui, 1999; Hutter & Legg, 2006), the application of niching methods (Mahfoud, 1995; Das et al., 2011), and the use of subpopulations (Ursem, 1999; Dezinger & Kidney, 2003). They are all designed to prevent population being trapped in one location. In contrast, other search methods, such as memetic algorithm (MA), are built on the assumption that EA provide significant diversity. This implies that landscape exploration is driven by the EA, and the exploitation of promising regions is left to local search methods (Molina et al., 2010). In reality, the explorative ability of MA is often only implicitly addressed. As a result, the performance of MA and other previously presented strategies is commonly evaluated indirectly by comparing their results (best fitness or average fitness) with those of other algorithms that do not implement the proposed features (Ursem, 2002). For instance, the performance of niching methods is frequently measured based on the number of peaks identified (Sareni & Krähenbühl, 1998). Of course, this technique is limited to problems having known optima locations. A more appropriate way to evaluate the performance of these strategies would be direct assessment. For this, the use of a diversity measure is preferable, since it allows for better characterization of the search behavior, and so provides a framework for algorithm comparison. Furthermore, tracking the diversity history throughout the process would make it possible to manage the exploration/exploitation balance (EEB) often sought by EA parameter control strategies (Lobo, Lima, & Michalewicz, 2007).

Two types of measurement are convenient for diversity monitoring: the Genotypic Diversity Measure (GDM), which characterizes the distribution of a population over a landscape, and the Phenotypic Diversity Measure (PDM), which describes the fitness distribution (Herrera & Lozano, 1996). GDM is more reliable than PDM for tracing premature convergence issues and for comparing the performance of multi-site convergence search processes, since the latter is influenced by the landscape relief. However, it is more difficult to assess diversity with GDM than it is with PDM, given that GDM is built on a multivariate distribution instead of a univariate distribution, as is the case for PDM (Tirronen & Neri, 2009).

---

[1] Corresponding author.
*E-mail addresses:* guillaume.eng.corriveau@gmail.com  (G. Corriveau), raynald.guilbault@etsmtl.ca  (R. Guilbault), antoine.tahan@etsmtl.ca (A. Tahan), robert.sabourin@etsmtl.ca (R. Sabourin).

In spite of the inherent complexity of GDM, numerous formulations have been proposed in the literature for the real-coded representation context. They can be classified into the following two families: distance-based measures, and gene frequency measures. The distance-based measurements (D) consider the distance between individuals, which can be evaluated from the mean spatial position of the population like the distance-to-average-point measure ($D_{DTAP}$) (Ursem, 2002; Abbass & Deb, 2003), the moment of inertia measure ($D_{MI}$) (Morrison & De Jong, 2002) or from the position of the fittest individual ($D_{ED}$) (Herrera & Lozano, 1996). The position of each individual could also be used. This evaluation ranges from the pairwise measure ($D_{PW}$) (Olorunda & Engelbrecht, 2008; Barker & Martin, 2000) to the maximum distance between two individuals ($D_{RP}$) (Olorunda & Engelbrecht, 2008). The second family scans the gene frequency (GF). This concept is generalized from binary representation, where the probability of the alleles at each locus is calculated within the entire population (Wineberg & Oppacher, 2003). In contrast, for a real-coded framework, all genes are continuous. Consequently, the gene scanning operation requires gene partitioning, where predefined intervals are considered as possible alleles. The number of intervals ($M$) involved in the discretization constitutes a severe limitation, as this number directly influences diversity estimation, especially for small populations or high dimensionality problems. Moreover, the gene frequency combination among all the landscape variables must be defined. For instance, in (Gouvêa Jr. & Araújo, 2008), a representative gene was preferred over averaging the diversity contribution of each gene (Wineberg & Oppacher, 2003).

Having many definitions of the same measure raises the question, what are the qualities of a good GDM? Table 1 lists three recognized quality criteria (Olorunda & Engelbrecht, 2008; Corriveau et al., 2012) that are desirable for a diversity indicator. Assessment frameworks are also proposed in the table. It is difficult to rank these criteria in terms of desirability, and so we consider them all to be equally important.

**Table 1**
Desirable GDM quality criteria

| QUALITY CRITERION | ASSESSMENT FRAMEWORK |
| --- | --- |
| Repeatability | Measuring the variance or the dispersion range of GDM values within the repeated evolution process |
| Robustness | Friedman statistical test: <br> a) with different population size ($N$) as sample <br> b) with different landscape dimensionality ($n$) as sample |
| Outlier handling capability | Measuring the diversity differences between a population without any outliers and a population with a fraction of the individuals acting as outliers |

Olorunda and Engelbrecht (2008) compare six GDMs on four test functions treated with a particle swarm optimization (PSO) approach. They rank the diversity measures according to their sensitivity to outliers. In contrast, Wineberg and Oppacher (2003) show that variance-based diversity measures, as well as the gene frequency family, are variants of the same basic concept: the sum of the distance between all possible pairs of elements considered. They conclude that experiments are not required for selecting the best GDM. However, in a recent study, Corriveau *et al.* (2012) present very different conclusions. The authors compare 15 GDMs defined within real-coded representation based on the criteria in Table 1. Their results show that the mean pairwise distance between the individuals in the population ($D_{PW}$) yields better diversity descriptions than other GDMs. Nevertheless, the response of $D_{PW}$ is inadequate when convergence appears over multiple locations.

This leads to the question of whether or not $D_{PW}$ and the other distance-based measures are capable of describing population diversity efficiently. If they are not, then the quality criteria in Table 1 would seem to be insufficient for appropriate diversity measure selection, and the following question has to be answered: *Do any available GDMs truly reflect population diversity?*

To the best of the authors' knowledge, no framework is available in the literature to validate the capability of a GDM as a diversity monitoring indicator. Not only must such a framework be provided, but a reliable GDM formulation must be identified to ensure accurate description of search behavior. This research addresses these challenging issues.

The paper is organized as follows: the next section provides the background of our GDM validation study; section 3 introduces diversity requirements for GDM validation purposes; section 4 describes the behavior of typical GDMs with respect to the proposed validation framework; and section 5 presents our concluding discussion.

## 2. Problem statement

The following simulations illustrate the response of two GDMs ($D^N_{PW}$ and $GF^N_S$). These results were obtained from a generic benchmark (Corriveau et al., 2012) which simulates the convergence process of a population over single-site and multi-site locations. This process is depicted in Fig. 1, where the rectangles represent the hyperspace allowed to the individuals associated with a given optimum. This generic benchmark does not account for the fitness distribution. Instead, the optima are randomly defined over the landscape at the beginning of the process. The hyperspaces shrink over a 51 iteration schedule, until all the individuals pile up on their respective optimum. The proposed convergence is simulated without any genetic operator, and the individuals are randomly generated at each iteration within their hyperspace boundaries. This generic benchmark eliminates any search bias coming from the operator. The simulations presented were conducted with a population of 100 individuals over a two-dimensional landscape.

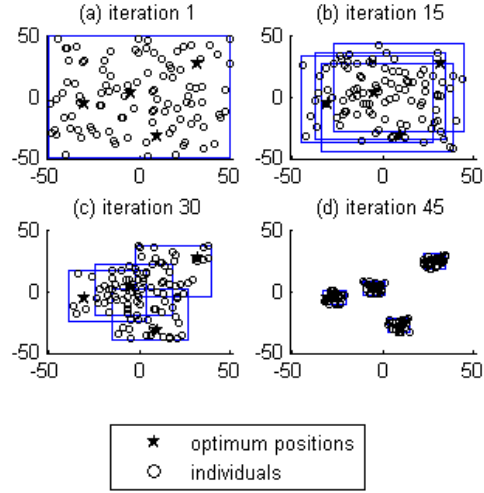

**Fig. 2.** Generic benchmark proposed in (Corriveau et al., 2012) with a population of individuals uniformly distributed at random (N=100) on four optimum positions in a two-dimensional landscape at given iterations.

Fig. 2 presents the diversity mean value history for 50 repetitions with the normalized version of $D_{PW}$ ($D^N_{PW}$). The normalization is based on the maximum diversity achieved so far in the optimization process (*NMDF*) (Corriveau et al., 2012). Fig. 2 also includes the normalized Shannon entropy ($GF^N_S$) (Shannon, 1948), which is a recognized gene frequency measurement. The normalization of $GF_S$ is realized with its maximum value. This is achieved when the gene frequencies are similar over the gene intervals ($M$), which means that the fraction of the population ($p_{m,k}$) belonging to interval $m \in \{1,\ldots, M\}$ on gene $k \in \{1,\ldots,n\}$ must equal $1/M$, where $n$ stands for the landscape dimensionality. In this experiment, $M$ was set to 100. However, it is important to note that this is true only if $M \leq N$, where $N$ represents the population size. Otherwise, the maximum value is obtained when $p_{m,k} = 1/N$. The formulation of $D_{PW}$ and $GF_S$ is given by (1) and (2) respectively. The notation used throughout this paper is presented in Table 2.

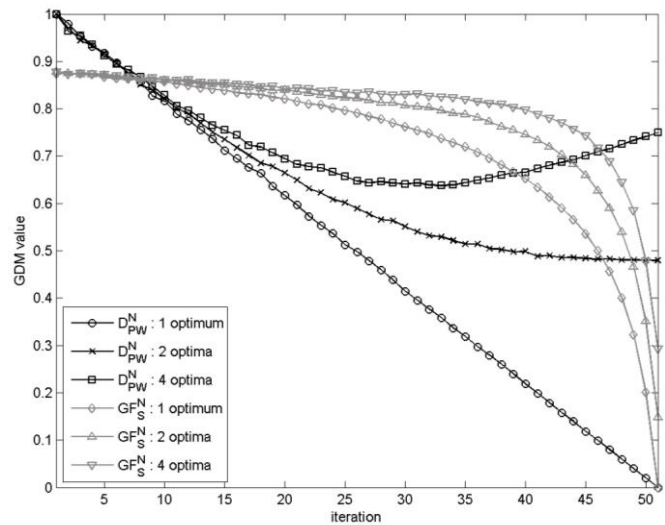

**Fig. 1.** Genotypic diversity levels of $D^N_{PW}$ and $GF^N_S$ over the single-site and multi-site convergence processes (two and four optima).

$$D_{PW} = \frac{2}{N(N-1)} \sum_{i=2}^{N} \sum_{j=1}^{i-1} \sqrt{\sum_{k=1}^{n} \left( x_{i,k} - x_{j,k} \right)^2} \tag{1}$$

$$GF_S(M) = -\frac{1}{n} \sum_{k=1}^{n} \sum_{m=1}^{M} p_{m,k} \log\left( p_{m,k} \right) \tag{2}$$

In (1), $x_{i,k}$ and $x_{j,k}$ represent the location of gene $k \in \{1,\ldots, n\}$ of the individual $i$ and $j \in \{1,\ldots, N\}$ respectively.

As indicated in Fig. 2, the $D^N_{PW}$ end diversity estimations are 48% and 75% for the two- and four-optima landscapes respectively. This obviously represents an overestimation of the true population diversity, since the final population (iteration 51) is concentrated at two/four sites. This overestimation behavior results from the deficient treatment of duplicate individuals in $D^N_{PW}$ (Ulrich, Bader & Thiele, 2010), as in other distance-based GDMs (Corriveau et al., 2012; Lacevic, Konjicija & Avdagic, 2007). In contrast, $GF^N_S$ seems to better describe the end diversity at convergence for multi-site processes. Nevertheless, as can be seen in Fig. 2, $GF^N_S$ does not offer representative diversity discrimination during the process, even for single-site convergence. This is explained by the fact that all $GF$ measures are based on the proportion of individuals resident in the various intervals for each gene, and the location of these intervals over the gene axis is not considered (Corriveau et al., 2012). In other words, the diversity variations become obvious only when most of the individuals pile up in the same interval.

**Table 2**
Definition of variables.

| VARIABLE | DEFINITION |
| --- | --- |
| $d(x_i, x_j)$ | Distance function used to monitor diversity between individuals $i$ and $j$ |
| $D(P)$ | Diversity of population $P$ |
| $E$ | Set of pairwise connections between individuals |
| $G(X, E)$ | Set of undirected graph represented by $X$ and $E$ |
| $i, j$ | Individual number $\in \{1, 2, \ldots, N\}$ |
| $k$ | Gene locus $\in \{1, 2, \ldots, n\}$ |
| $l$ | Length of the side of an hypercube $S$ |
| $m$ | Interval number |
| $M$ | Total number of intervals |
| $n$ | Landscape dimensionality |
| $N$ | Population size |
| $p_{m,k}$ | Fraction of $N$ that belong to interval $m$ on gene $k$ |
| $P$ | Population |
| $S(x_i, l)$ | Hypercube bounding the diversity contribution of individual $i$ |
| $U(P)$ | Population $P$ based on a uniform distribution |
| $V$ | Volume of the landscape |
| $x_{i,k}$ | Gene $k$ of individual $i$ |
| $\hat{x}_i$ | Normalized location of individual $i$ |
| $X$ | Set of individuals location |
| $LB_k$ | Lower bound of gene $k$ |
| $MST$ | Minimum spanning tree |
| $NMDF$ | Normalization with maximum diversity so far |
| $UB_k$ | Upper bound of gene $k$ |
| $\mu$ | Total length of the $MST$ |
| $\mu_L(A)$ | Lebesgue measure of a set $A$ |

The previous observations indicate that none of these GDMs seems to be valuable over the multi-site convergence process. This makes assessing the underlying performance of any diversity promoting strategy troublesome. Moreover, even for search algorithms converging intentionally or not to single-site, duplicate individuals are always a possibility throughout the evolution process. Consequently, any population-based search process may suffer from diversity distortion and so mislead the search behavior analysis.

In response to the weakness of the previous indicators, Lacevic, Konjicija, and Avdagic (2007) proposed the volume-based measure (L-diversity) as the GDM. They argued that it is probably the most intuitive and accurate way to evaluate diversity of a population. This measure is designed to compute the volume of the union of $n$ axis-aligned hyper-rectangles. In computational geometry, this is known as the Klee measure problem (KMP) (Klee, 1977), and it represents a generalization of the dominated hypervolume measure used in multi-objective optimization problems (MOOP) for assessing the approximation quality of the Pareto front (Beume & Rudolph, 2006). The L-diversity, referred to here as $D_L$, is given by:

$$D_L = \mu_L\left(\bigcup_{i=1}^{N} S(\boldsymbol{x}_i, l)\right) \tag{3}$$

where $\mu_L(A)$ represents the Lebesgue measure of a set A. The parameter $l$ corresponds to the length of the side of a hypercube $S(\boldsymbol{x}_i, l)$ bounding the diversity contribution of the individual $\boldsymbol{x}_i$. Setting $l = \sqrt[n]{V/N}$ promotes full coverage of the search space volume ($V$) when the individuals are uniformly distributed. $D_L$ suffers from its computational complexity exponentially growing with respect to the dimensionality of the landscape, leading to $O(N^{\lfloor n/2 \rfloor})$ when all the hypercubes have the same size (Boissonnat et al., 1995). This condition makes $D_L$ practically intractable as GDM.

This problem led Lacevic, Konjicija, and Avdagic (2007) to searching which measure best approximates $D_L$. They based their investigation on a correlation analysis over various controlled population arrangements. This study was later extended to include more GDMs in (Lacevic & Amaldi, 2011). As a result, the Euclidean minimum spanning tree measure ($D_{MST}$) turns out to be the best alternative to $D_L$. Its formulation is defined by:

$$D_{MST} = \mu\left(MST\big(G(X,E)\big)\right) \tag{4}$$

where $MST(G(X,E))$ represents the minimum spanning tree subgraph of the complete undirected graph $G(X,E)$, which is defined by the set $X$ representing the location of the individuals of the population and the set of edges $E$ denoting all the pairwise connections between individuals. The summation of the total length of the $MST$ subgraph is symbolized by $\mu$. The rationale behind the $D_{MST}$ proposal is to extract only the "principal" distances, in order to alleviate the issue of duplicated individuals (Lacevic & Amaldi, 2011). Fig. 3 illustrates a 2D example of the diversity evaluation mechanism of $D_L$ and $D_{MST}$.

In addition, Lacevic and Amaldi (2011) developed the theoretical concept of ectropy for evaluating to what extent an indicator penalizes duplicate individuals. The ectropy concept helped justify the use of $D_L$ as a reference in the correlation study; the maximum evaluation of $D_L$ is never obtained in presence of duplicate individuals. Ectropy was also used for illustrating the weakness of $D_{PW}$ and other distance-based measurements. However, ectropy analysis was restricted to a limited set of GDMs, due to the difficulty of analytically deriving the maximal state of any formulation. This analysis illustrates the limited capacities of theoretical development in assessing the relevance of GDMs.

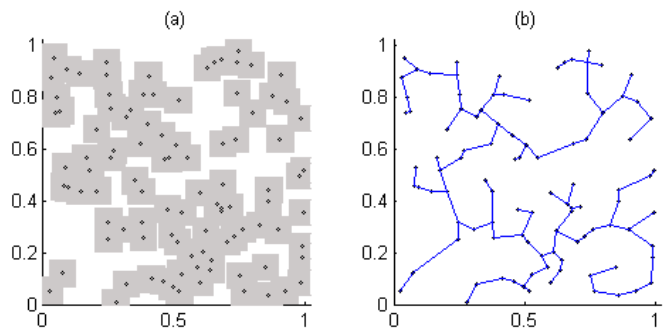

**Fig. 3.** Representation of a uniformly random population with 100 individuals bounded between $[0, 1]^2$, where diversity is evaluated by: (a) $D_L$ – union of the area associated with each individual, (b) $D_{MST}$ – total length of the MST.

### 3. Characterizing population diversity

As mentioned, numerous GDM formulations are available as well as different analysis frameworks for their comparison. However, the lack of precise feature characterizing population diversity makes the choice of the best measure problematic. Defining such requirements may provide common ground for validating which GDM accurately describes population diversities.

In pioneering research, Weitzman (1992) listed fourteen salient characteristics of reliable measures. Among them, six are considered to be mathematical characteristics, two are categorized as taxonomic aspects, one is an ecological consideration, and five are economic considerations. The Weitzman properties are summarized in Table 3. Weitzman acknowledged that these properties are not equally important. Later, Solow and Polasky (1994) identified three of them as fundamental requirements:

1.  Monotonicity in species: adding a species (an individual, in the current context) should not decrease the diversity $D$, or $D(P') \leq D(P)$, if $P'$ is a subset of population $P$.

2.  Twinning: the addition of an individual or a species already in the population should not increase the diversity $D$, or $D(P \cup x_i) = D(P)$, if the distance between individuals $x_i$ and $x_j$, $d(x_i, x_j) = 0$, where $x_j \in P$ and $x_i \notin P$.

3.  Monotonicity in distance: an unambiguous increase in distance between individuals should be reflected in the diversity measurement $D$, or $D(P') \leq D(P)$. This requirement reflects the following situation, all the elements in population $P$ equal those in population $P'$, except individuals $x_i$ and $x_j$ from population $P$, and $x_i'$ and $x_j'$ from population $P'$, while their distances respect the following inequality: $d(x_i', x_j') \leq d(x_i, x_j)$.

**Table 3**
Diversity properties defined by Weitzman (1992).

| # | PROPERTY | CATEGORY |
|---|----------|----------|
| 1 | Monotonicity in species | Mathematical |
| 2 | Link property | Mathematical |
| 3 | Twin property | Mathematical |
| 4 | Continuity in distance | Mathematical |
| 5 | Monotonicity in distance | Mathematical |
| 6 | Maximum diversity that can be added by a species | Mathematical |
| 7 | Clade aggregation | Taxonomic |
| 8 | Ultrametric distances reduce diversity theory to perfect taxonomy theory | Taxonomic |
| 9 | Removal of false diversities by identifying individuals from the same species set | Ecological |
| 10 | Favor the more distantly related species | Economic |
| 11 | Irrelevance of equally distant relatives | Economic |
| 12 | Rule of the snake | Economic |
| 13 | Additivity properties of induced utility functions | Economic |
| 14 | Min-loss extinction | Economic |

Even though the diversity measures studied by Weitzman (1992) and Solow & Polasky (1994) were not formulated for the present context, the proposed fundamental requirements are still suitable for evaluating GDM trueness in reflecting a diversity measure. In reality, diversity measurement should be understood as a coverage space indicator. This concept is completely and rigorously expressed by those diversity requirements. Therefore, the three requirements are adapted to EA real-coded GDMs in Table 4.

Monotonicity in species will be referred to here as *monotonicity in individual varieties*. This is a more general expression, and is applicable in the EA context, since maximal diversity is achieved with a uniformly distributed population ($U(P)$). Such a population is constructed by ensuring that, on each gene, individuals are separated by the same distance. This distance is defined by $(UB_k\text{-}LB_k)/(N^{1/n}\text{-}1)$, where $LB_k$ and $UB_k$ represent the lower and upper bounds of the landscape $k$ axis respectively. This requirement establishes the upper bound of the possible diversity of a population. The mathematical formulation becomes $D\left(P'\right) \leq D\left(P\right) \leq D\left(U\left(P\right)\right)$, where $P'$ is a subset of population $P$.

The initial definition of the *twinning* requirement is directly transferrable to the present context. However, for fixed population sizes, the existence of duplicate individuals inevitably reduces the diversity of a population. The mathematical form becomes $D((P \setminus x_q) \cup x_i) \leq D(P)$ if $d(x_i, x_j) = 0$, where $x_j \in P$, $x_i \notin P$, and $x_q$ is an individual removed from the population $P$ to make room for $x_i$. As a matter of fact, the twinning requirement has the same meaning as the ectropy concept described before.

Finally, the requirement of monotonicity in distance is reformulated to highlight the fact that genotypic diversity should be based on the location of the various individuals. For example, considering two uniformly distributed populations ($U(P_A)$ and $U(P_B)$) over region A and B of the same landscape, the corresponding diversities should present the following relation: $D\left(U\left(P_A\right)\right) < D\left(U\left(P_B\right)\right)$, if

$$\prod_{k=1}^{n}\left(UB_k - LB_k\right)_A < \prod_{k=1}^{n}\left(UB_k - LB_k\right)_B .$$

**Table 4**
Defined requirements for GDM trueness validation.

| # | REQUIREMENT | BRIEF DESCRIPTION |
|---|---|---|
| 1 | Monotonicity in individual varieties | - Adding a non-duplicate individual should not decrease diversity<br>- A uniformly distributed population provides upper bound diversity |
| 2 | Twinning | Duplicate individuals should reduce diversity as the population moves away from a uniformly distributed population |
| 3 | Monotonicity in distances | Diversity should decrease as individuals move closer together |

## 4. Validation of the representative GDMs

In this section, only representative GDMs are considered. Therefore $D_{PW}$ and $GF_{S_x}$ with $M = 10$, are selected to characterize common distance-based and gene frequency measurements respectively, while $D_L$ and $D_{MST}$ are included as potential GDM candidates following the recommendation in (Lacevic & Amaldi, 2011). The validation analyses the response of the GDMs to three diversity requirements on two frameworks: a reduced population arrangement, and various controlled cases of population diversity, as explained below.

### 4.1 Reduced population arrangement

The first framework intends to validate the general behavior of the GDMs in a simple an intuitive manner. A population of 5 individuals ($P_5$) is promoted on a 2D landscape bounded between $[0, 1]^2$, four of these individuals are fixed at the landscape corner ($x_1 = (0, 0)$, $x_2 = (1, 0)$, $x_3 = (1, 1)$, $x_4 = (0, 1)$), and the remaining individual ($x_5$) is moved on the diagonal connecting $x_1$ and $x_3$. This framework makes it possible to break down the multivariate aspect of GDM into a univariate problem by tracking the diversity variation of the normalized location of $x_5$ ($\hat{x}_5$). For comparative purposes, diversity of a static population with 4 individuals ($P_4$) located at $x_1$ to $x_4$ is also included.

To respect the diversity requirements established in Table 4, the following conditions must be satisfied:

1. Monotonicity in individual varieties: $D\left(P_5 \middle| \hat{x}_5 = 0.5\right) \geq D\left(P_4\right)$;

2. Twinning: $\min D\left(P_5\right) = D\left(P_5 \middle| \hat{x}_5 = 0 \vee \hat{x}_5 = 1\right)$;

3. Monotonicity in distance: $\max D(P_5) = D\left(P_5 \middle| \hat{x}_5 = 0.5\right)$.

The results of this framework are presented in Fig. 4. The charts indicate that $D_{PW}$ is unable to respect any of the diversity requirements, as $D(P_5)$ is always lower than $D(P_4)$ and the maximum diversity state of $P_5$ is achieved in the presence of duplicate individuals ($\hat{x}_5 = 0 \wedge \hat{x}_5 = 1$). A similar conclusion may be drawn for $D_{MST}$. It is interesting to note, however, that $D_{MST}$ gives the same diversity for $P_5$ with $x_5$ at boundaries (duplicate individuals) than for $P_4$. This is obvious from the *MST* computation standpoint, but it demonstrates that $D_{MST}$ has a problem penalizing duplicate individuals. In fact, this issue stems from the disagreement between the summation of the "principal" distances and the monotonicity in distance. In other words, the diversity level of $D_{MST}$ with $P_5$ is neither monotonic nor decreasing, as $x_5$ comes closer to $x_1$ or $x_3$. As a matter of fact, the local peaks obtained by $D_{MST}$ around $\hat{x}_s = 0.3$ and $\hat{x}_s = 0.7$ are due to changes in the MST connections.

On this reduced population framework, $GF_S$ and $D_L$ show good respect of all three diversity requirements. Clearly, diversity level discrimination is better for $D_L$ than for $GF_S$. This is in accordance with the issue described in section 2 on gene frequency measurements. Furthermore, the evaluation of $D_L$, $D\left(P_5 \mid \hat{x}_s = 0.5\right) = D\left(P_4\right)$, is more likely to conform with the population size robustness criterion (Table 1). Having the upper bound of the diversity included in the $D_L$ formulation, through the definition of $l$, makes the measurement independent of the population size parameter.

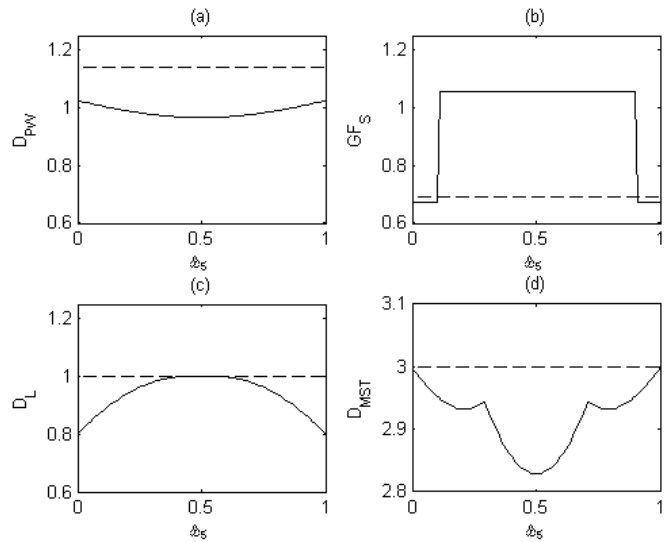

**Fig. 4.** Diversity on $P_5$ (solid curves) and $P_4$ (dash curves) with respect to the normalized location of $x_S$ evaluated from: (a) $D_{PW}$. (b) $G_{FS}$. (c) $D_L$. (d) $D_{MST}$.

### 4.2    Controlled cases of population diversity

The second GDM trueness validation framework involves the examination of seven frozen cases of population diversity. Besides the difference in the population arrangements, the benefit of this framework is a better representation of common EA population sizes. A population size ($N$) of 100 is used for all cases on a 2D landscape bounded between $[-1, 1]^2$. These simple deterministic cases allow us to illustrate the three requirements listed in Table 4, while at the same time avoiding costly simulations. Of the seven cases, which are defined below, four are directly related to the modality of the landscape (individuals attached to predefined optima (Cases 2 to 5)).

Case 1:    The population is fixed at one point on the landscape.
Case 2:    The population is distributed with a uniform ratio on the optima located at a mid-point between the landscape center and corners.
Case 3:    The population is distributed with a non-uniform ratio on the optima located at a mid-point between the landscape center and corners.
Case 4:    The population is distributed with a uniform ratio on the optima located at the corners of the landscape.
Case 5:    The population is distributed with a non-uniform ratio on the optima located at the corners of the landscape.
Case 6:    The population is distributed uniformly over the landscape diagonal.
Case 7:    The population is distributed uniformly over the landscape.

Case 1 and 7 simulate the complete convergence and full diversity conditions of a genotypic population respectively. Cases 2 and 3 and Cases 4 and 5 offer an identical geographical position. However, in Cases 3 and 5, one optimum monopolizes 70% of the individuals, with the rest equally distributed over the remaining optima. Fig. 5 presents the geographical map of the population for these cases. To validate the coherence of GDMs response over multi-site locations, a two- and four- optima landscape are considered for these four cases.

Therefore, for Cases 3 and 5 with two-optima, the individuals match the 70/30 arrangement, while for the four-optima landscape, the individuals follow a 70/10/10/10 distribution. Case 6 corresponds to a situation where an individual would only have identical gene values, with those values evenly spaced among individuals. This is described by $x_{i,k} = LB_k + (i-1)*(UB_k - LB_k)/(N-1)$, $\forall$ $k \in \{1,\dots, n\}$, where $i \in \{1,\dots, N\}$. In such a situation, the individuals would be distributed along a landscape diagonal.

The frozen case list is presented in increasing order of diversity. Consequently, to respect the 1st requirement in Table 4, an adequate GDM will rank the cases in the same order. Moreover, considering the twinning, Case 2 will be equal to Case 3, and Case 4 will be equal to Case 5. Finally, the monotonicity in distance is accounted for if Cases 4 and 5 present higher diversities than Cases 2 and 3. Table 5 presents the results obtained for all cases with $D_{PW}$, $GF_S$, $D_L$, and $D_{MST}$. Within this framework, the interval number ($M$) used by $GF_S$ is set to 100.

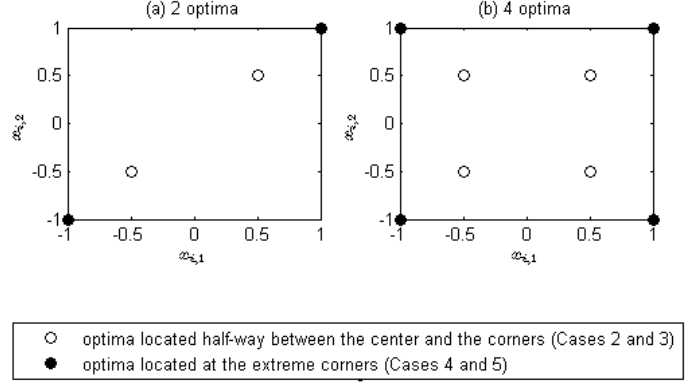

**Fig. 5.** Position of the optima for Cases 2 to 5 on: (a) 2-optima landscape, and (b) 4 optima landscape.

Since the diversity levels obtained are higher for the cases where the individuals are located at the landscape corners (Cases 4 and 5) than for Case 7, Table 5 indicates that $D_{PW}$ does not respect the 1st requirement. In addition, the diversity estimations for Cases 2 and 4 are higher than for Cases 3 and 5 respectively. This reveals the additional contribution of the duplicate individuals within $D_{PW}$, which indicates that the 2nd requirement is not respected either. In addition, since Case 6 exhibits a lower diversity than Cases 4 and 5, Table 5 also reveals that $D_{PW}$ does not fulfill the requirement of monotonicity in distance. Based on these observations, the frozen case experiment accurately reflects the observed shortcoming of distance-based measures over the multi-site convergence process (section 2).

Table 5 also indicates that $GF_S$ violates all three requirements. Diversity assessment by the aggregation of each gene leads to violation of the 1st requirement, since considering each gene independently increases the diagonal distribution (Case 6) diversity estimation. This observation demonstrates that the generalization by aggregation of a univariate diversity indicator into a multivariate framework can be problematic. The 2nd requirement is not respected, since the distribution of the duplicate individuals impacts the diversity level (Case 2 ≠ Case 3, and Case 4 ≠ Case 5). Finally, the diversity level does not decrease as the optima move closer to one another (Case 2 ≮ Case 4, and Case 3 ≮ Case 5), and so the 3rd requirement is not respected either.

The results of Table 5 indicate that $D_L$ cannot respect the requirement of monotonicity in distance; because $D_L$

**Table 5**
Behavior of the representative GDMs over the seven frozen cases.

| GDM | LANDSCAPE | GENOTYPIC DISTRIBUTION CASES | | | | | | |
|---|---|---|---|---|---|---|---|---|
| | | 1 | 2 | 3 | 4 | 5 | 6 | 7 |
| $D_{PW}$ | 2 optima | 0 | 0.71 | 0.60 | 1.43 | 1.20 | 0.96 | 1.16 |
| | 4 optima | | 0.86 | 0.55 | 1.72 | 1.10 | | |
| $GF_S$ | 2 optima | 0 | 0.69 | 0.61 | 0.69 | 0.61 | 4.61 | 2.30 |
| | 4 optima | | 0.69 | 0.50 | 0.69 | 0.50 | | |
| $D_L$ | 2 optima | 0.04 | 0.08 | 0.08 | 0.08 | 0.08 | 0.80 | 4.00 |
| | 4 optima | | 0.16 | 0.16 | 0.16 | 0.16 | | |
| $D_{MST}$ | 2 optima | 0 | 1.41 | 1.41 | 2.83 | 2.83 | 2.83 | 22.00 |
| | 4 optima | | 3.00 | 3.00 | 6.00 | 6.00 | | |

aggregates the volume covered by each individual regardless of their locations, the descriptor makes no difference between the optima location (Case 2 = Case 4, and Case 3 = Case 5).

Finally, Table 5 indicates that $D_{MST}$ violates the requirement of monotonicity in distance, and to some extent that of monotonicity in individual varieties. In fact, no distinction appears between cases with all individuals fixed at the corner (Case 4 and Case 5) and cases with individuals set on the landscape diagonal (Case 6). These cases share the same MST, although the diversity state of Case 6 is higher than that of Cases 4 and 5.

*4.3    Discussion*

Table 6 summarizes the results obtained from the two frameworks. The superscripts indicate the framework revealing the deficient response. The aggregation of these results demonstrates that the two frameworks, taken individually, are insufficient for a complete validation of GDMs. On the other hand, associated, they offer efficient validation of GDM performances. In addition, Table 6 particularly reveals that none of the studied GDM guarantees accurate description of the population diversity. We are therefore forced to conclude that all evaluated measurements could represent a misleading factor in monitoring diversity.

**Table 6**
Summary of the fulfillment of the diversity requirements by the representative GDMs (A − violation identified through the reduced population arrangement framework. B − violation identified through the controlled cases of population diversity framework).

| GDM | REQUIREMENTS | | |
|---|---|---|---|
| | **1**<br>**Monotonicity in individual varieties** | **2**<br>**Twinning** | **3**<br>**Monotonicity in distance** |
| $D_{PW}$ | NO[A,B] | NO[A,B] | NO[A,B] |
| $GF_S$ | NO[B] | NO[B] | NO[B] |
| $D_L$ | YES | YES | NO[B] |
| $D_{MST}$ | NO[A,B] | NO[A] | NO[A,B] |

## 5.    Conclusion

Genotypic diversity measurement (GDM) is a useful concept for monitoring and/or managing the exploration of an optimization process. Premature convergence towards a suboptimal solution can be minimized through strategies using the information gathered by a GDM. Multiple GDMs have been proposed in the literature over the years. However, to the best of the authors' knowledge, their ability to describe population diversity has never been exhaustively investigated. In GDM-related applications as well as in GDMs comparison study, the assumption that a particular GDM truly reflects population diversity is often adopted. However, the issues observed with some of these formulations, such as poor handling of duplicate individuals, lead us to question the trustworthiness of this premise. Consequently, using a GDM not fulfilling this assumption can potentially disrupt the analysis of the search process.

The aim of this paper is to look at the development of a framework that allows GDM to be assessed as population diversity descriptors. To achieve this, we extracted three diversity requirements from the literature to form the basis for our investigation. The requirements are: monotonicity in individual varieties, twinning, and monotonicity in distance. These diversity requirements are intuitive properties that GDM must have, in order to offer an accurate coverage space description. Our study here is restricted to real-coded representation, although the established diversity requirements are not limited to this context. We identified and evaluated four GDMs from previous studies: the mean pairwise measure ($D_{PW}$), the Shannon entropy ($GF_S$), the L-diversity or volume-based measure ($D_L$), and the minimum spanning tree measure ($D_{MST}$).

The response of the selected GDMs to the requirements was evaluated by means of two validation frameworks involving a reduced population arrangement of 4 and 5 individuals, and seven test cases with controlled population diversity. These simple frameworks showed that the three diversity requirements are sufficient for proper evaluation of the GDM response. The frameworks also served to identify and characterize the limitations of the available GDMs.

In summary, $D_{PW}$, $GF_S$, and $D_{MST}$ showed improper response to all three diversity requirements. Mostly because they do not consider a uniformly distributed population as the most diverse state. They also present some difficulties in managing duplicate individuals and cannot efficiently account for relative locations of the individuals within the population. On the other hand, $D_L$ was revealed to be the sole formulation able to meet two of the three requirements. Nevertheless, besides its prohibitive computational cost, it offers no reliable mechanism to account for the requirement of monotonicity in distance. As illustrated by the controlled cases of population diversity framework, its failure to meet the third requirement could impact the diversity analysis when the population is configured in non intersecting clusters.

Beside these limitations that must be overcome to develop an adequate GDM, we caution as a general recommendation that care should be exercised regarding the generalization of a univariate diversity indicator by aggregation into a multivariate context for GDM purposes. In addition, special attention should be paid to monotonicity in distance during the development of new formulations, since no tested GDM was able to completely meet this diversity requirement.

In this investigation, no theoretical development was included for the characterization of the GDMs as population diversity indicators. The objective behind this decision was to assure general evaluations and provide a generic framework supporting any kind of GDM formulations. The obtained results showed the benefit of the chosen approach; the investigated GDM responses are described by means of practical requirements. Nevertheless, since they are needed to help the introduction of suitable GDMs, in particular able to cope with the monotonicity in distance, theoretical developments supporting the practical requirements represent the next step.

Globally, the present investigation demonstrated that the definition of an adequate genotypic diversity formulation for real-coded representation remains an open question. Moreover, the proposed GDM validation frameworks will facilitate the evaluation of any new proposals, by relating simple cases of controlled diversity to the fundamental requirements that the diversity descriptor must exhibit. It is important to mention that even if the proposed GDM validation framework combination was sufficient for detection of inadequate response of the tested GDMs, the reciprocal should not be assumed: the framework combination alone remains insufficient to guarantee the validity of a given GDM. The proposed evaluation tool should only be considered as a first gate, since the GDM must be tested, thereafter, within higher dimensionality landscapes.